\documentclass[mlmain]{jmlr}

\usepackage{longtable}

\usepackage{booktabs}
\usepackage[load-configurations=version-1]{siunitx} 

\theorembodyfont{\upshape}
\theoremheaderfont{\scshape}
\theorempostheader{:}
\theoremsep{\newline}

\jmlryear{2025}
\jmlrworkshop{Symmetry and Geometry in Neural Representations}

\title[Generalizable World Models via VSA]{Geometric Priors for Generalizable World Models \titlebreak via Vector Symbolic Architecture}

 \author{
\Name{William Youngwoo Chung} \Email{chungwy1@uci.edu}\\
\addr University of California, Irvine, United States
\AND
\Name{Calvin Yeung} \Email{chyeung2@uci.edu}\\
\addr University of California, Irvine, United States
\AND
\Name{Hansen Jin Lillemark} \Email{hlillemark@ucsd.edu}\\
\addr University of California, San Diego, United States
\AND
\Name{Zhuowen Zou} \Email{zhuowez1@uci.edu}\\
\addr University of California, Irvine, United States
\AND
\Name{Xiangjian Liu} \Email{xiangjl4@uci.edu}\\
\addr University of California, Irvine, United States
\AND
\Name{Mohsen Imani} \Email{m.imani@uci.edu}\\
\addr University of California, Irvine, United States
}

\begin{document}

\maketitle

\begin{abstract}
A key challenge in artificial intelligence and neuroscience is understanding how neural systems learn representations that capture the underlying dynamics of the world. Most world models represent the transition function with unstructured neural networks, limiting interpretability, sample efficiency, and generalization to unseen states or action compositions. We address these issues with a generalizable world model grounded in \textit{Vector Symbolic Architecture} (VSA) principles as geometric priors. Our approach utilizes learnable Fourier Holographic Reduced Representation (FHRR) encoders to map states and actions into a high-dimensional complex vector space with learned group structure and models transitions with element-wise complex multiplication. We formalize the framework’s group-theoretic foundation and show how training such structured representations to be approximately invariant enables strong multi-step composition directly in latent space and generalization performances over various experiments. On a discrete grid world environment, our model achieves 87.5\% zero-shot accuracy to unseen state-action pairs, obtains 53.6\% higher accuracy on 20-timestep horizon rollouts, and demonstrates $4\times$ higher robustness to noise relative to an MLP baseline. These results highlight how training to have latent group structure yields generalizable, data-efficient, and interpretable world models, providing a principled pathway toward structured models for real-world planning and reasoning.
\end{abstract}

\section{Introduction}
\label{sec:intro}
Humans build internal world models that capture the underlying dynamics of the environment and allow interaction beyond direct trial-and-error \citep{lecun2022path}. Inspired by this idea, modern reinforcement learning (RL) has leveraged latent predictive models conditioned on states and actions, achieving state-of-the-art results in video games~\citep{hafner2020mastering} and continuous control~\citep{hansen2023td}. Despite these successes, current world models face two major limitations. They are largely confined to RL, control, and robotics settings where abundant simulated data is available and treat the transition function $T: S \times A \to S$ as an \textbf{unstructured} black-box function approximators. While highly expressive, such architectures suffer from poor sample efficiency, weak extrapolation to unseen states, compounding rollout errors, and latent spaces with no explicit geometric meaning. 

Biological systems, in contrast, appear to exploit \textit{symmetries} and \textit{geometric structure}~\citep{gardner2022toroidal, gallego2017neural} in the environment, effectively reducing the complexity of learning. Geometric Deep Learning (GDL)~\citep{bronstein2021geometric, Papillon_2025, shewmake2023visual} formalizes this idea by incorporating geometric priors into neural networks to help preserve structure throughout the network, dramatically improving the efficiency and generalization capabilities.

Vector Symbolic Architecture (VSA)~\citep{kleyko2022survey} offers a complementary, algebraic approach to structured representations. They represent symbols as high-dimensional vectors and compose them via binary operations, forming structured representations that are robust to noise. In addition, VSA-based representations can be trained to approximate group actions, making them a promising approach for GDL. Among the many VSA variants, Fourier Holographic Reduced Representation (FHRR)~\citep{plate2003holographic} remains a popular implementation of VSA to efficiently encode complex data structures due to its efficiency and exact invertibility.
\vspace{-0.5em}
\paragraph{Contribution.} In this work, we propose a generalizable world model using VSA principles. Our FHRR encoder encodes states and actions as unitary complex vectors, with transitions realized as element-wise multiplication. We train the model to have latent group structure on actions with multi-step action composition, invertibility, and robust cleanup. We demonstrate that the model (1) encourages transition equivariance in the latent space and learns action representations respecting group structure, (2) achieves long-horizon stability and error correction via cleanup, and (3) outperforms MLP baselines on one-step prediction, long-horizon rollouts, zero-shot generalization, and robustness tests on a discrete grid world environment regardless of scale. Our approach provides an alternative architecture to world modeling with strong implications for interpretable and generalizable decision-making for real world applications.

\section{Related Work}
\label{sec:related_work}
\paragraph{World Models in Model-Based RL.}\label{sec:worldmodels}
World models aim to learn a predictive model of an environment's transition dynamics that can leveraged for learning and planning. Model-based reinforcement learning methods~\citep{ha2018world, hafner2019dream} and recent Model Predictive Control methods~\citep{hansen2023td} utilize world models to learn the environment's dynamics for decision-making, yet often suffer from compounding rollout errors~\citep{hansen2022temporal} and limited transparency in the learned dynamics~\citep{glanois2024survey}. Most approaches treat the transition function as an unstructured mapping from $(s,a) \mapsto s'$, which can be highly expressive but fails to exploit known symmetries in the environment. This can limit generalization, especially when the environment exhibits strong symmetries or if little training data is available. These limitations motivate the incorporation of \textit{geometric priors} into world models, where known symmetries or structures in the environment are part of the representation learning process.
\vspace{-0.5em}
\paragraph{Geometric Deep Learning.}\label{sec:gdl}
In the context of world modeling, GDL-based approaches \citep{kipf2019contrastive, park2022learning} have incorporated symmetry and structure to improve generalization in structured environments. However, in such implementations, the latent representations themselves are not structured such that they can be algebraically composed, inverted, or directly manipulated. Without the ability to easily or interpretably manipulate latents with vector operations, planning or composition with such architectures consequently require an expensive full forward pass or additionally trained modules. Our VSA-based approach to GDL trains the action representations such that it respects group structure in the latent space and leverages cleanup for robustness and compounding error reduction.
\vspace{-0.5em}
\paragraph{Vector Symbolic Architectures.}\label{sec:vsa}
VSA, also known as Hyperdimensional Computing, is a computational paradigm where discrete symbols are represented as high-dimensional vectors (e.g. $D \geq 1000$) sampled from well-defined distributions \citep{kanerva2009hyperdimensional}. These representations are inherently robust to noise and enable symbolic reasoning through simple algebraic operations. VSAs have been applied in diverse domains, including efficient classification~\citep{ni2024heal}, time-series modeling~\citep{mejri2024novel}, graph reasoning~\citep{poduval2022graphd}, reinforcement learning~\citep{ni2024efficient}, and representing cognitive maps~\citep{yeungCognitiveMapFormation2025}. Hardware-efficient implementations have also made them appealing for resource-constrained applications and learning on the edge~\citep{zou2021spiking, chung2025robust}. However its applications as a transition operation in learnable settings remain largely unexplored. Our work aims to close this gap by constructing \textit{learnable} world models that utilize the binding operation to model environment transitions and the VSA clean-up mechanism to perform robust rollouts. 
\begin{figure}[htbp]
    \centering
    \includegraphics[width=1\linewidth]{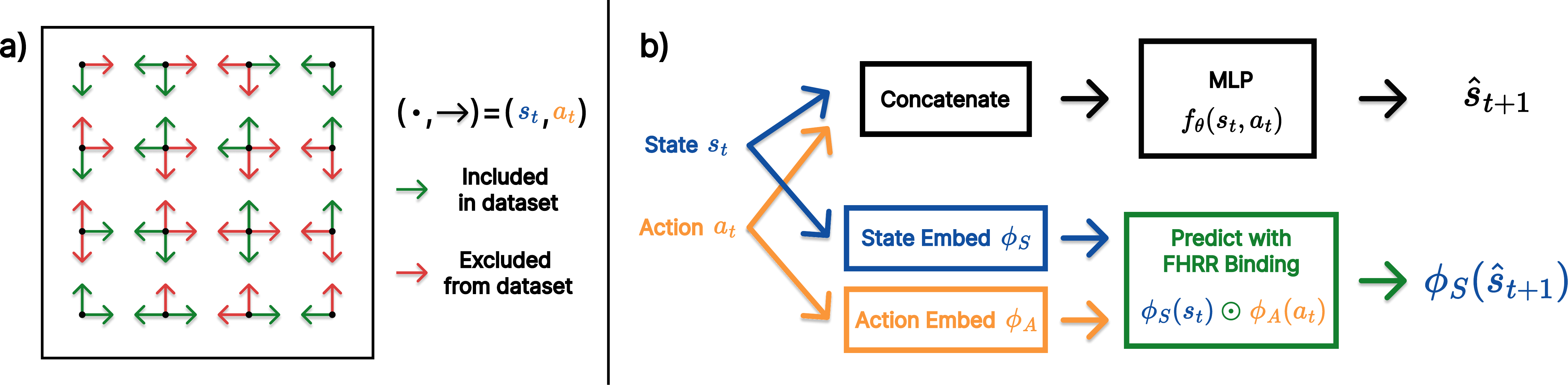}
    \caption{Overview of the FHRR-based world modeling framework. a) Visualization of the partially held-out GridWorld Environment. b) Difference between MLP-based and FHRR-based dynamics modeling. Direct predictions of $\hat{s}_{t+1}$ by MLPs cannot easily generalize to OOD samples, while FHRR can.}
    \label{fig:framework}
    \vspace{-3em}
\end{figure}
\section{Proposed Framework}
\label{sec:method}
\figureref{fig:framework} visualizes the architecture of our generalizable world model. We model environment dynamics as group actions in a learnable, complex latent space where the dynamics is implemented by the binding operation. Concretely, we learn state and action encoders such that the transition model inherits several useful VSA properties including interpretability of the transformation between two given states, robustness to noise due to the high-dimensionality of the representations, and a native cleanup mechanism to mitigate exponential error growth in long-horizon rollouts. In contrast, MLP-based approaches concatenates of state-action pairs which limits the model's ability to learn separate state and action mappings.  

\subsection{Fourier Holographic Reduced Representation}

FHRR~\citep{plate2003holographic} is a specific VSA variant in which each vector component lies on the unit circle in the complex plane, i.e.
\begin{equation}
    \mathbf{v} = [e^{i\theta_j}]_{j=1}^D\in\mathbb{C}^D
\end{equation}
such that $\theta_{j} \sim p$ for dimension $j=1,\dots,D$ where $p$ is some distribution (e.g. $\mathrm{Unif}(0,2\pi)$ or $\mathcal{N}(0,1)$), resulting in a phase-based complex unitary vector. As a VSA, FHRR admits two binary operations, namely bundling ($+$) and binding ($\odot$), to construct composite representations. Bundling is implemented as vector addition while binding is implemented as element-wise complex multiplication. Given vectors $\mathbf{v}_1 = [e^{i\theta_{1,j}}]_{j=1}^D$ and $\mathbf{v}_2 = [e^{i\theta_{2,j}}]_{j=1}^D$, $\mathbf{v}_1\odot \mathbf{v}_2=[e^{i(\theta_{1,j}+\theta_{2,j})}]_{j=1}^D$. The inverse of a FHRR vector \(\mathbf{v}\) is simply its complex conjugate, enabling straightforward unbinding via \(\mathbf{v}^{-1} = \overline{\mathbf{v}}\). A notable property of FHRR is its connection to kernel-based methods in machine learning (\appendixref{apd:kernel}). There are various other related implementations of VSA (\appendixref{apd:vsa}), but we limit the scope of VSA to FHRR in this work.

\subsection{Environment Dynamics as a Group Action on Sets}
\label{sec:env_group_action}
\vspace{-0.5em} \noindent
Let $S$ be a finite set of states, $A$ a finite set of actions, and $T : S \times A \to S$ a deterministic transition function. We assume that compositions of actions generate an \emph{action group} $(G, \circ)$ acting on $S$:
\begin{equation}
\cdot : G \times S \to S, 
\quad (g, s) \mapsto g \cdot s,
\end{equation}
with identity $e \cdot s = s$ and $(g_1 \circ g_2) \cdot s = g_1 \cdot (g_2 \cdot s)$ for $g_1,g_2\in G$. Each action $a \in A$ corresponds to a generator $g_a \in G$ such that $T(s, a) = g_a \cdot s$. In particular, the identity $e$ does not need to be a primitive action, but is always present in $G$ and can be realized as $e = g_a \circ g_a^{-1}$ for any $a \in A$.
 

\subsection{Equivariant Latent Representations}
\label{sec:equiv_hom}
\vspace{-0.5em} \noindent
We embed states into a $D$-dimensional complex vector space via a map $\phi_S:S\to\mathcal{Z}$, where $\mathcal{Z} = (S^1)^D = \{ z \in \mathbb{C}^D \ :\ |z_d| = 1 \}$. Notably, $(\mathcal{Z},\odot)$ forms a group. A representation of the action group $G$ in $\mathcal{Z}$ is a \textit{homomorphism}:
\begin{equation}
\rho : G \to \mathcal{Z}, \quad 
\rho(g_1 \circ g_2) = \rho(g_1) \odot \rho(g_2), \quad \forall g_1, g_2 \in G.
\end{equation}

\noindent The encoder $\phi_S : S \to \mathcal{Z}$ is \emph{equivariant} to environment transitions if
\begin{equation}
\label{eq:equivariance}
\phi_S(T(s,a)) = \rho(a) \odot \phi_S(s), \quad \forall s \in S, \ a \in A
\end{equation}
where $T(s,a)$ is the environment's transition function. By closure of $G$, the same property holds for any composed action $g \in G$, i.e. $\phi_S(g \cdot s) = \rho(g) \odot \phi_S(s)$, which states that transforming $s$ by $g$ corresponds to multiplying its latent representation by $\rho(g)$.

\subsection{Latent Transition Model}
\label{sec:latent_transition}
\vspace{-0.5em} \noindent
We would like to learn state and action encoders, $\phi_S:S\to \mathcal{Z}$ and $\phi_A:A\to\mathcal{Z}$ respectively, such that (1) $\phi_A$ induces a representation of $G$ via the generators $g_a\in G$ for all $a\in A$; and (2) the equivariance condition given by Eq.~\ref{eq:equivariance} holds. 

Suppose $s\in\mathbb{R}^{n_s}$ and $a\in\mathbb{R}^{n_a}$ where $n_s$ and $n_a$ are the original state and action dimensions, respectively. We parameterize $\phi_S$ and $\phi_A$ via the FHRR encoding:
\begin{align}
    \phi_S(s)&=[e^{i\theta_{j,s}^\top s}]_{j=1}^D,\quad\phi_A(a)=[e^{i\theta_{j,a}^\top a}]_{j=1}^D.
\end{align}
Motivated by Eq.~\ref{eq:equivariance}, we model the latent transitions in FHRR-space via the binding operator
\begin{align}
    \tau:\mathcal{Z}\times \mathcal{Z}&\to \mathcal{Z},\quad(\phi_S(s),\phi_A(a))\mapsto \phi_S(s)\odot \phi_A(a)
\end{align}
We would also like to learn $\phi_S:S\to\mathcal{Z}$ and $\phi_A:A\to\mathcal{Z}$ such that one-step dynamics satisfy
\begin{align}
\label{eq:one_step}
\phi_S(s_{t+1}) &= \tau(\phi_S(s_{t}),\phi_A(a_t))= \phi_S(s_t)\odot \phi_A(a_t),
\end{align}
and $\phi_A(a)=\rho(g_a)$. In phase coordinates, this corresponds to:
\begin{align}
    \Theta_{s}^\top{s_{t+1}} = \Theta_s^\top {s_{t}} + \Theta_a^\top {a_{t}} \ \ (\mathrm{mod} \ 2\pi)
\end{align}
where $\Theta_s=[\theta_{j,s}]_{j=1}^D\in\mathbb{C}^{D\times n_s}$, $\Theta_a=[\theta_{j,a}]_{j=1}^D\in\mathbb{C}^{D\times n_a}$, and the modulus is applied element-wise. Due to the properties of FHRR, we can simply extend this to multi-step composition:
\begin{align}
\textbf{Embedding Space: }&\quad\phi_S(s_{t+k})=\phi_S(s_t)\odot \prod_{j=1}^k \phi_A(a_{t+j-1})\label{rollout}\\
\textbf{Phase Space: }&\quad\Theta_s^\top {s_{t+k}} = \Theta_s^\top{s_t}+ \sum_{j=1}^k \Theta_a^\top{a_{t+j-1}} \quad (\mathrm{mod} \ 2\pi).\label{eq:fast_rollout}
\end{align}

\subsection{Learning Objectives}
\label{sec:learning}
\vspace{-0.5em} \noindent
We learn $\phi_S$ and $\phi_A$ with learnable parameters $\Theta_s$ and $\Theta_a$ respectively and train on transition tuples, $(s_t,a_t,s_{t+1})$. We minimize a binding loss to encourage transition equivariance given in Eq.~\ref{eq:one_step}:
\begin{equation}
\label{binding_loss}
\mathcal{L}_{\text{bind}} = \| \phi_S(s_{t+1})-\phi_S(s_t)\odot\phi_A(a_t)\|^2
\end{equation}
Additionally, to preserve structure in our representations, we introduce invertibility and orthogonality regularizers:
\begin{equation}\label{eq:inv}
 \mathcal{L}_{\text{inv}} = \sum_{(a, a^{-1})} \left\| \phi_A(a) \odot \phi_A(a^{-1}) - \mathbf{1} \right\|^2,
\end{equation}
\begin{equation}
\mathcal{L}_{\text{ortho}} = \sum_{i \neq j} \left( \langle \phi_S(s_i), \phi_S(s_j) \rangle \right)^2.
\end{equation}
In particular, the invertability constraint given by Eq.~\ref{eq:inv} encourages that the actions $a\in A$ form a representation via $\phi_A$, i.e. $\phi_A$ induces an approximate homomorphism with respect to the actions in the grid environment.

The full objective is $\mathcal{L}=\lambda_{\text{bind}}\mathcal{L}_{\text{bind}}+\lambda_{\text{inv}}\mathcal{L}_{\text{inv}}+\lambda_{\text{ortho}}\mathcal{L}_{\text{ortho}}$ where \( \lambda_{\text{bind}}, \lambda_{\text{inv}}, \lambda_{\text{ortho}} \) are the hyperparameters controlling the balance between each objective respectively. Training is linear-time in $D$ per sample and memory is $O(D)$ as all VSA operations are done element-wise. Additionally, multi-step rollouts can be processed using Eq.~\ref{eq:fast_rollout} for linear-time in the phase space where $(|a| \in A) \ll D$ for inference.

\begin{figure}[htbp]
    \centering
    \includegraphics[width=\linewidth]{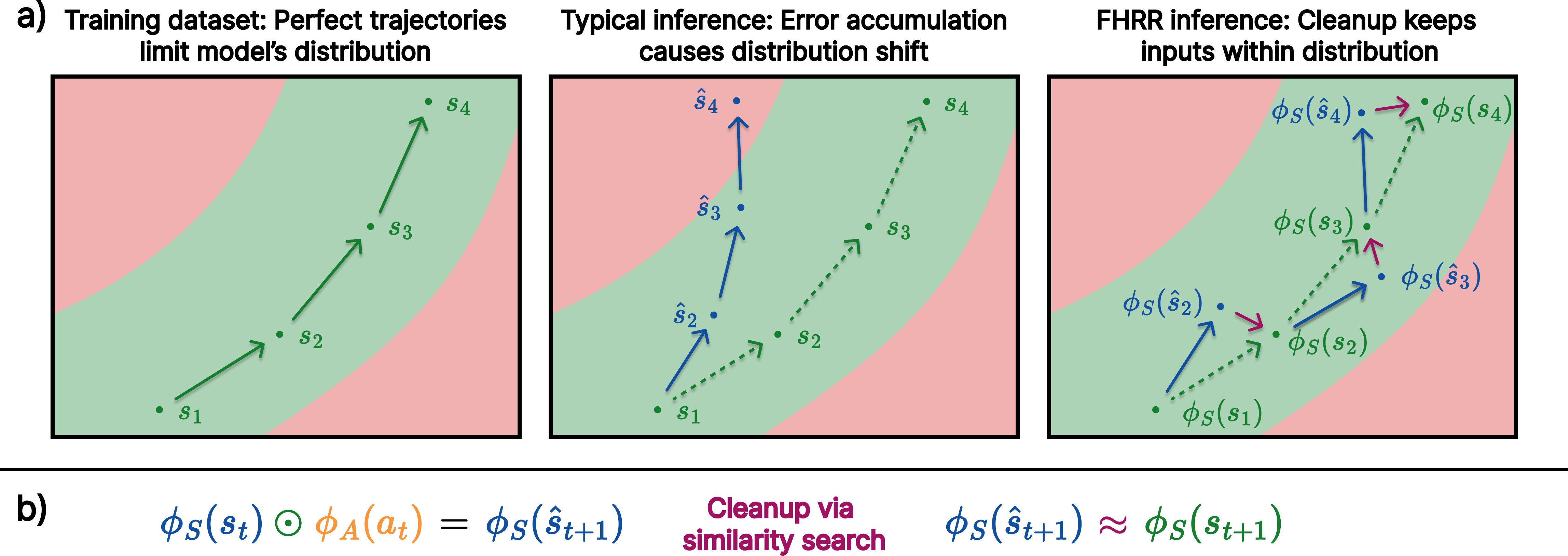}
    \caption{a) Shaded red represents out of the training distribution. Cleanup during FHRR Inference can ameliorate error accumulation due to distribution shift between train and test. b) Cleanup mechanism equations via similarity search in FHRR.}
    \label{fig:cleanup}
    \vspace{-6mm}
\end{figure}

\subsection{Cleanup Mechanism}
\label{sec:cleanup}
\vspace{-0.5em} \noindent
A key advantage of VSA-based models is their ability to support self-correct (error correction) through a process known as \emph{cleanup}~\cite{plate1991holographic}, visualized in Figure \ref{fig:cleanup}. Given a noisy or approximate prediction of the next-state, $\phi_S(\hat{s}_{t+1})$, cleanup recovers the most likely true embedding by similarity search i.e. $\phi_S(\hat{s}_{t+1}) = \arg\max_{s \in \mathcal{S}} \ \mathrm{Re} \,\langle \phi_S(\hat{s}_{t+1}), \phi_S(s) \rangle.$ where $\mathrm{Re}\langle \quad \rangle$ refers to taking the real part of the complex inner product between two vectors.

Cleanup is based on the premise that in high-dimensional spaces, randomly generated vectors are almost always far apart from one another. As a result, a noisy vector still remains noticeably closer to its true state embedding than to any other state. Intuitively, one can visualize it as the level of separation between random vectors increases as the dimensionality increases. 

In our setting, each state $s \in S$ is encoded as a unitary complex vector, and during training the \(\mathcal{L}_{\text{ortho}}\) term (Section~\ref{sec:learning}) encourages that distinct state representations are \emph{quasi-orthogonal}, i.e. $\mathrm{Re} \,\langle \phi_S(s), \phi_S(s') \rangle \approx 0 \quad \text{for } s \neq s'.$ We maintain a \textit{state codebook}, whose rows store the learned embeddings for every learned discrete state in the environment. During inference an approximate prediction of the next state is cleaned up by simply comparing this prediction to the entries in the codebook and selecting the most similar one. This process works because different state embeddings are trained to be nearly orthogonal, producing large separation margins in high dimensions. More formal descriptions of the cleanup mechanism are provided in Appendix~\ref{apd:cleanup}.

\begin{table}[hbtp]
\vspace{-5mm}
\setlength{\tabcolsep}{3pt}
\label{tbl:HDC_ACC_Results}
\floatconts
  {tab:example-booktabs}
  {\caption{VSA-Based vs MLP Dynamics Modeling}}
  {\begin{tabular}{lccccc}
  \toprule
  Task &  FHRR (Ours) & MLP-Small & MLP-Medium & MLP-Large \\
  \midrule
1-step Accuracy & \textbf{96.3\%} & 80.0\% & 80.0\% & 80.25\% \\
1-step Accuracy (Zero-Shot) & \textbf{87.5\%} & 0.0\% & 0.0\% & 1.25\% \\
Cosine Similarity & \textbf{83.0} & 79.5 & 79.9 & 80.6 \\
Cosine Similarity (Zero-Shot) & \textbf{80.5} & 0.9 & 0.15 & 3.1 \\
Rollout (5 steps) & \textbf{74.6\%} & 39.8\% & 38.0\% & 40.8\% \\
Rollout (20 steps) & \textbf{34.6\%} & 2.0\% & 4.0\% & 6.2\% \\
Rollout (20 steps + Clean) & \textbf{61.4\%} & 5.4\% & 7.8\% & 8.4\% \\
Rollout (100 steps) & 1.8\% & 0.8\% & 1.8\% & \textbf{2.0\%} \\
Rollout (100 steps + Clean) & \textbf{38.6\%} & 2.8\% & 4.0\% & 3.2\% \\
  \bottomrule
  \end{tabular}}
\end{table}
\vspace{-5mm}

\section{Results}
\label{sec:experiments}

\paragraph{Experimental Design.} We train and evaluate our VSA-based world models with 3 MLP baselines of varying sizes on a 10\texttimes10 GridWorld environment with a total of 100 discrete states and 4 deterministic actions. For all models, we train on 80\% of (state, action) pairs, hold out 20\% for zero-shot evaluation, and train for 500 epochs. In~\tableref{tbl:HDC_ACC_Results}, we compare FHRR versus MLP-Small, MLP-Medium, and MLP-Large on several different tasks, such as 1-step accuracy, cosine similarity, and rollouts. In~\appendixref{apd:inference}, we compare the total parameter count and inference time between all the models to highlight that VSA models have a similar parameter count as MLP-Small. For more explicit details on the implementation, please check the~\appendixref{apd:implementation}.

\begin{figure}[htbp]
\vspace{-2mm}
    \centering
    \includegraphics[width=0.9\linewidth]{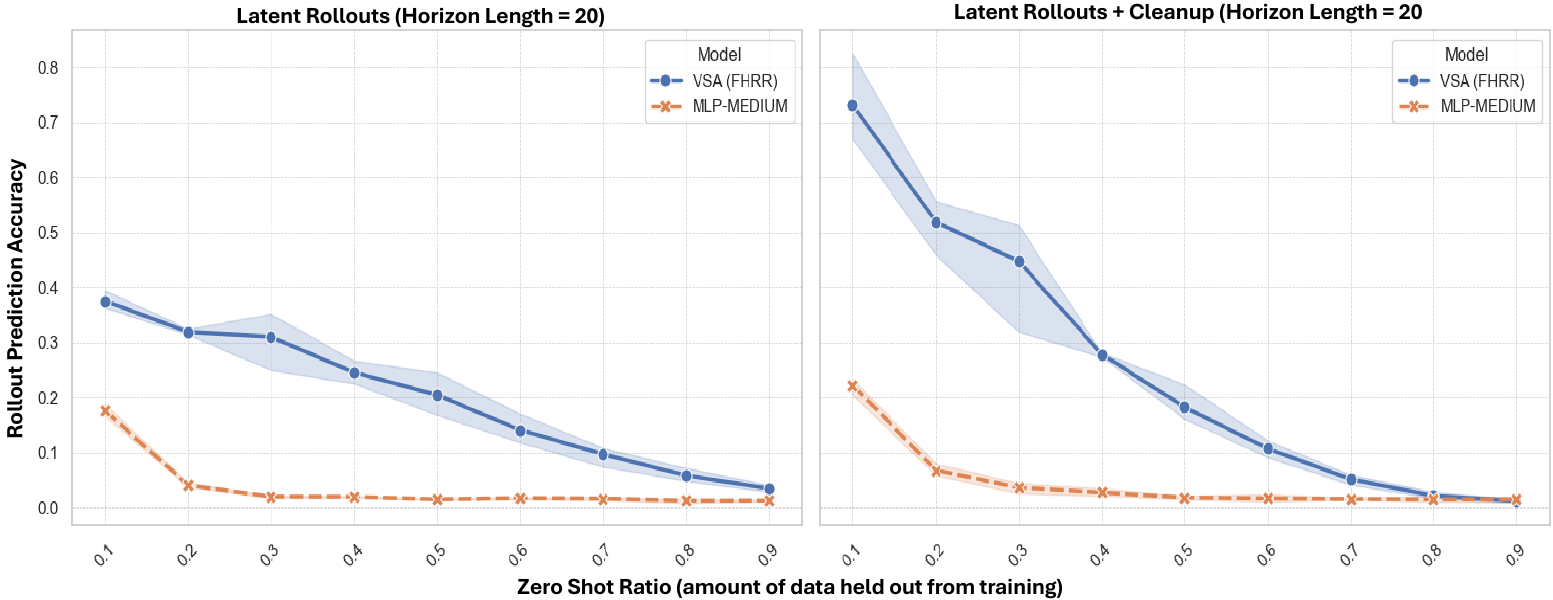}
    \caption{Latent Rollout Accuracy of FHRR and MLP-M over varying zero-shot ratios}
    \label{fig:zeroshot}
\vspace{-2mm}
\end{figure}

\paragraph{Dynamics Modeling} For 1-step accuracy, we test the models on their ability to predict the correct next state given (state, action) pairs. While the models are trained on 80\% of the dataset, they are tested on all possible transitions. 
For the zero-shot tests (when evaluating on the unseen 20\% transitions), our model achieves significantly higher zero-shot accuracy and cosine similarity, confirming its ability to generalize well. We also highlight that scaling the size of the MLPs has not shown a significantly stronger ability for these models to generalize. In the rollout tests, (i.e. interactions solely in the latent space, FHRR maintains a higher accuracy over long transitions, unlike MLP which accumulates drift. Additionally, VSA has the added benefit of applying \textit{cleanup}. For a fair comparison, we utilize nearest-neighbor search for the MLPs to compare against the VSA model with cleanup and apply the cleanup every 2 time steps.

\begin{figure}
    \vspace{-2mm}
    \centering
    \includegraphics[width=0.9\linewidth]{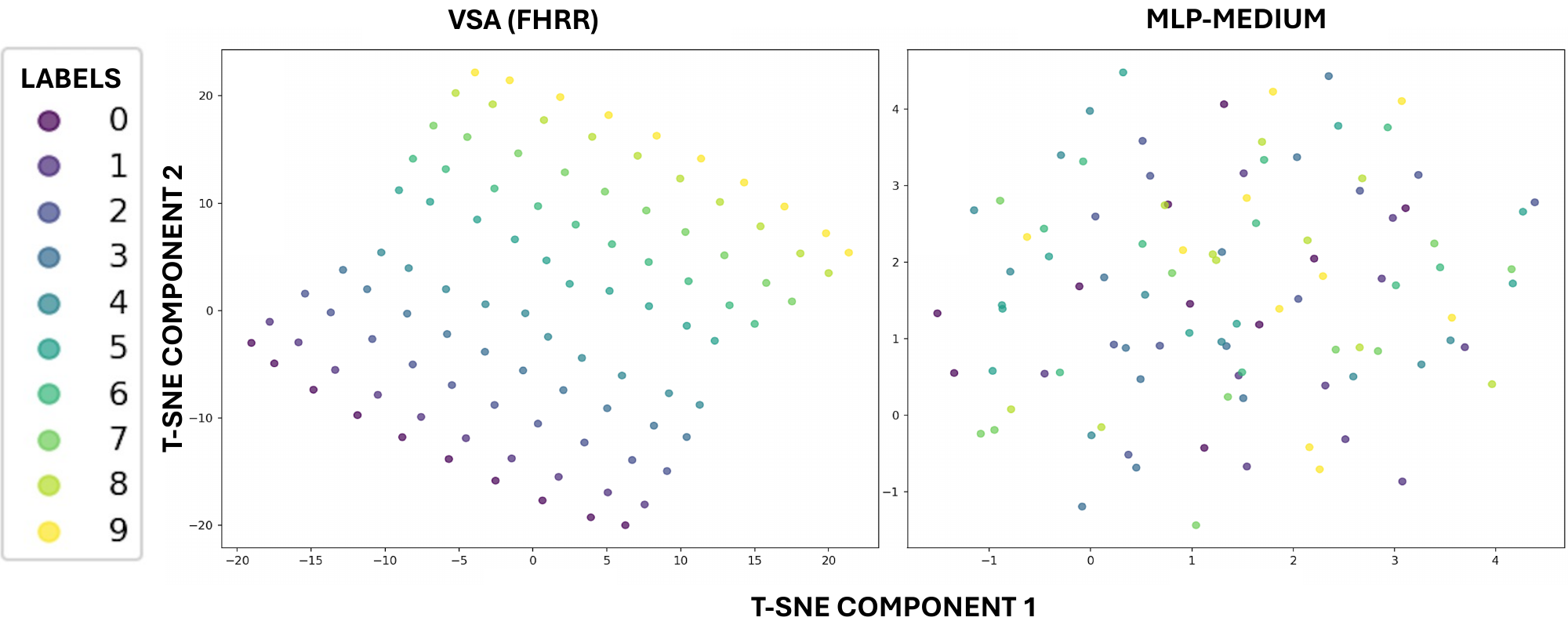}
    \caption{t-SNE visualization of state embeddings for VSA (FHRR) vs MLP-M labeled per row}
    \label{fig:latent_vis}
    \vspace{-3mm}
\end{figure}

\paragraph{Latent Rollout Performance.} 
\figureref{fig:zeroshot} compares the latent rollout accuracy of horizon length $t = 20$ between FHRR and MLP-Medium given varying zero-shot ratios. As the zero-shot ratio increases, we notice a linear decrease in the FHRR model performance while the MLP-based model's accuracy exponentially decreases and fails to maintain even above 10\% when trained on only 90\% of the transitions. Additionally, the cleanup operation improves the accuracy of the FHRR model by 35\% when zero-shot ratio = 0.1, resulting in a $3.3\times$ improvement over the MLP baseline.

\paragraph{Latent Visualizations.} In~\figureref{fig:latent_vis} we visualize the t-SNE components of the state embeddings of the FHRR and MLP-Medium models. The FHRR model is able to capture the structure of the grid environment in its latent space while MLP-Medium fails to maintain any structure. We attribute this structured latent space for its strong generalization capabilities over MLP baselines.

\begin{figure}[htbp]
\vspace{-4mm}
\floatconts
  {fig:similarity_robustness}
  {\caption{FHRR vs MLP-Medium on robustness and similarity experiments}}
  {
    \subfigure[Robustness to Noise]{\label{fig:robustness}%
      \includegraphics[width=0.4\linewidth]{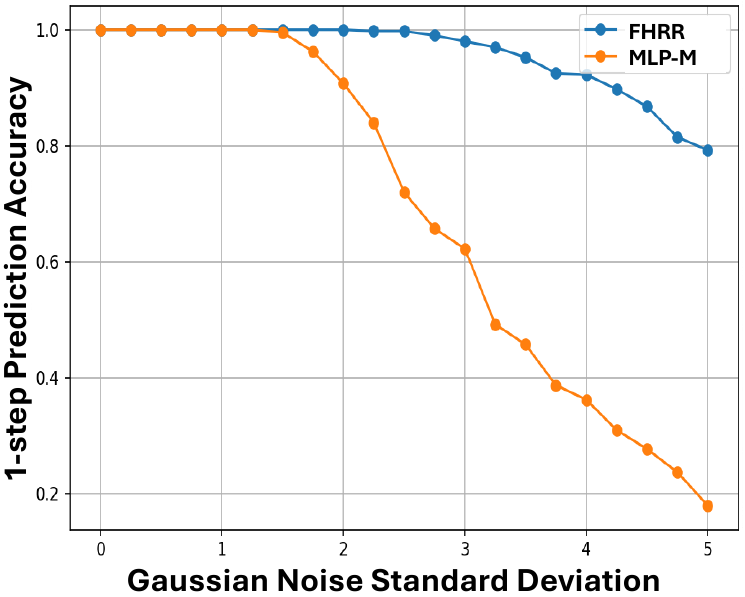}}%
    \qquad
    \subfigure[Similarity Kernels]{\label{fig:similarity}%
      \includegraphics[width=0.38\linewidth]{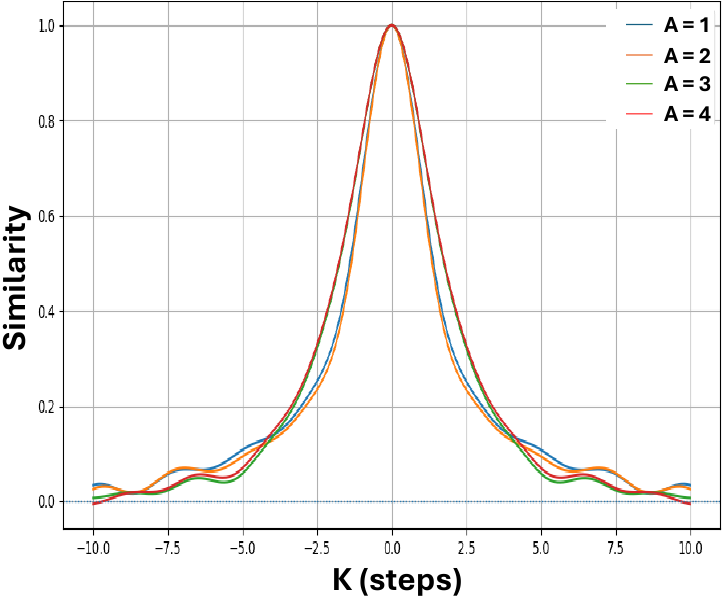}}
  }
\vspace{-2mm}
\end{figure}

\paragraph{Robustness.} We study the robustness of our FHRR model with MLP-Medium by comparing the 1-step dynamics accuracy when random gaussian noise is added to the transition function. \figureref{fig:robustness} shows the results between gaussian noise between 0 to 5 magnitude of standard deviation (i.e. noise $n \sim \mathcal{N}(0,\sigma) \text{ where $\sigma \in$[0, 5]}$). The FHRR model maintains above a 80\% accuracy even under large amounts of noise while MLP-M struggles as the scale of noise increases. 

\paragraph{Similarity Kernel.} Given the kernel approximation property of FHRR (\appendixref{apd:kernel}), in~\figureref{fig:similarity}, we plot the similarity between $\phi_S(s)$ and $\phi_S(s + ka)$ where $a$ is a given action and $k \in [-10, 10]$. For all actions, we notice an approximately smooth but sharp similarity kernel peaking at $k=0$ and decaying as $|k|$ increases, indicating that the latent space preserves locality of the states.
The approximate symmetry of these curves across actions demonstrates that our model learns a structured geometry in which actions correspond to consistent translation across the states.

\section{Conclusion}
\label{sec:conclusion}
In this work, we present an alternative approach to world modeling based on VSA, where states and actions are encoded as unitary complex vectors and transitions are modeled via element-wise multiplication. Our experiments on a discrete world environment show that our model achieves strong generalization capabilities and maintains long-horizon rollout accuracy under noise, outperforming MLP baselines regardless of scale. While promising, our current work is limited to small discrete environments. Extending this approach to continuous, stochastic, or partially observable domains remains an open challenge. In future work, we hope to integrate VSA-based world modeling into model-based RL and planning to enable generalizable dynamics models that are applicable to real world environments. Overall, this work demonstrates that learning structured algebraic representations offers a principled path toward robust, interpretable, and generalizable world models.

\section*{Acknowledgements}
This work was supported in part by Nthe DARPA Young Faculty Award, the National Science Foundation (NSF) under Grants \#2431561, \#2127780, \#2319198, \#2321840, \#2312517, and \#2235472, the Semiconductor Research Corporation (SRC), the Office of Naval Research through the Young Investigator Program Award, Grants \#N00014-21-1-2225 and \#N00014-22-1-2067, Army Research Office Grant \#W911NF2410360, and the National Defense Science \& Engineering Graduate (NDSEG) Fellowship Program. Additionally, support was provided by the Air Force Office of Scientific Research under Award \#FA9550-22-1-0253, along with generous gifts from Xilinx and Cisco. 

\bibliography{pmlr-sample}

\newpage

\appendix

\section{Vector Symbolic Architectures}\label{apd:vsa}

\subsection{VSA Variants}
VSA consists of two fundamental operations: \textbf{bundling} (superposition), which adds vectors to form set-like representations, and \textbf{binding} (association), which combines vectors into a compositional representation, often with invertibility. This algebraic structure allows VSAs to represent structured data such as sequences, sets, and relations in a way that supports compositionality and symbolic reasoning through lightweight vector operations. 

Many VSA works also consist of a \textbf{permutation} operation (shuffling) to encode ordered structures or design non-commutative operations when combined with \textbf{bundling} or \textbf{binding}. Below, we summarize some related VSA approaches:

\paragraph{Holographic Reduced Representations} 
HRR~\citep{plate1995holographic} uses real-valued vectors with components drawn from a normal distribution and defines binding as circular convolution, bundling as element-wise addition, and similarity as dot product between two vectors. Fourier Holographic Reduced Representation is the extension of Holographic Reduced Representations, which avoids convolution and fourier transforms altogether by using element-wise complex multiplication which is equivalent to convolution in the frequency domain~\citep{plate2003holographic}.

\paragraph{Multiply Add Permute} MAP~\citep{gaylerMultiplicativeBindingRepresentation1998} utilizes high-dimensional bipolar vectors where binding is element-wise multiplication, bundling is element-wise addition, and similarity as cosine similarity or the dot product. MAP forms the foundation for many subsequent VSA designs due to the simplicity of element-wise multiplication for binding.

\paragraph{Generalized Holographic Reduced Representations} 
GHRR~\citep{yeungGeneralizedHolographicReduced2024} is an extension of FHRR by replacing the unitary complex vector $e^{i\theta} \in U(1)$ with a unitary matrix $a_j \in U(m)$, such that vectors become tensors $H \in \mathbb{C}^{D \times m \times m}$ whose binding is defined as matrix multiplication. Bundling still holds as element-wise addition. The similarity between two hypervectors $H_1 = [a_j]_{j=1}^D$ and $H_2 = [b_j]_{j=1}^D$ is defined as
\begin{equation}
\delta(H_1, H_2) \;=\; \frac{1}{mD} \, \mathrm{Re} \left( \mathrm{tr} \sum_{j=1}^D a_j b_j^\dagger \right).
\end{equation}
For $m=1$, this reduces exactly to FHRR similarity. GHRR enables non-commutative and more flexible representations, controlled by the choice of unitary matrices yet equivalent to FHRR when $m=1$.

\subsection{Kernel Approximation in FHRR}\label{apd:kernel}
One way to encode data into FHRR follows the Random Fourier Features (RFF) encoding \citep{rahimi}, an efficient approximation of kernel methods. The RFF encoding is a map $\phi:\mathbb{R}^n\to\mathbb{C}^D$, with $\phi(\mathbf{x})=e^{i\mathbf{Mx}}$, where each row $\mathbf{M}_{j,:}\sim p$ for some multivariate distribution $p$. As a result of Bochner's theorem, $\langle\phi(\mathbf{x}),\phi(\mathbf{y})\rangle/D\approx K(\mathbf{x}-\mathbf{y})$ for all $\mathbf{x},\mathbf{y}\in\mathbb{R}^n$, where $K$ is a shift-invariant kernel that is the Fourier transform of distribution $p$. The approximation converges to the true kernel in the limit as $D\to\infty$. Notably, when $p$ is the standard Gaussian distribution, the radial basis function (RBF) kernel is recovered. 

\subsection{Cleanup in VSA}
\label{apd:cleanup}

An advantage of VSA representations is that different symbols (states in world modeling) are learned to be nearly orthogonal in high-dimensional space~\cite{plate1991holographic}. This means that small perturbations to a predicted state vector $\hat{s}_{t+1}$ typically do not change which symbol it is closest to, enabling reliable identity recovery through a simple nearest-neighbor search.

In contrast, conventional neural networks often learn entangled latent spaces with no explicit separation or compositional structure. In such spaces, even small prediction errors can move a representation across decision boundaries, causing semantic drift that accumulates over long rollouts. In settings involving hardware noise or approximate computation (e.g., in-memory or neuromorphic accelerators and world modeling) this becomes a larger issue. Given a perturbed output 
\[
\hat{s}_{t+1} = f_\theta(s, a) + \epsilon,
\]
there is no guarantee that
\[
\arg\min_{s'} \|f_\theta(s,a) - \hat{s}_{t+1}\| \approx s_{t+1},
\]
making nearest-neighbor lookup unreliable under large amounts of noise or long horizon prediction tasks.

\paragraph{Codebook-based cleanup.}
In VSA, cleanup is often implemented as a nearest-neighbor search over a \emph{state codebook}. Let $\Phi \in \mathbb{C}^{|\mathcal{S}| \times D}$ denote the matrix of state embeddings, where $\Phi_s = \phi_S(s)$. Given a noisy prediction $x$, cleanup selects the state with greatest real-part similarity:
\[
s^\star = \arg\max_{s \in \mathcal{S}} \ \mathrm{Re}\,\langle x, \Phi_s \rangle,
\]
which corresponds to taking the argmax over the real parts of the matrix multiplication between a state embedding and the state codebook. When the number of states is moderate (as in discrete environments), this cleanup incurs negligible overhead and provides identity-correcting feedback at every timestep. For larger $\mathcal{S}$, approximate nearest neighbor or restricted-batch cleanup can be used. When the state-codebook is explicitly stored as a matrix, this operation can be reduced to taking an $\arg\max$ after performing matrix-vector multiplication.

\paragraph{Why cleanup works: concentration in high dimensions.}
Two geometric properties guarantee the reliability of cleanup:

\emph{(1) Concentration of self-similarity.}  
If $x = \phi_S(s) + \eta$ is a noisy version of the true embedding, then
\[
\mathrm{Re}\,\langle x, \phi_S(s) \rangle 
\text{ concentrates around } 1,
\quad 
\operatorname{Var} = \mathcal{O}(1/D),
\]
so the effect of noise shrinks with dimension.

\emph{(2) Concentration of cross-similarities.}  
For any $s' \neq s$, quasi-orthogonality ensures
\[
\mathrm{Re}\,\langle \phi_S(s), \phi_S(s') \rangle \approx 0,
\quad
\operatorname{Var} = \mathcal{O}(1/D).
\]

Together, these imply a separation margin of \[
\mathrm{margin} \sim 1 - \mathcal{O}(1/\sqrt{D}),
\]
meaning that larger dimensionality produces exponentially more reliable cleanup.

\section{Implementation Details}
\label{apd:implementation}

\subsection{Environment}
\paragraph{Dataset}
We use a $10\times 10$ GridWorld with boundaries (no wrap-around). States are indexed and mapped to (row, col). 
Actions $a\in\{0,1,2,3\}$ correspond to \textit{up}, \textit{down}, \textit{left}, \textit{right} with deterministic transition $T(s,a) = (s')$.

We define transitions $(s,a,s')$ for $s\in S$, $a\in A$, which yields $|S||A|=100\times 4$ tuples. We form a zero-shot split at the level of given a zero-shot ratio, such that a fixed ratio (default $20\%$) of pairs $(s,a)$ are withheld from training. Zero-shot evaluations are done only held-out pairs, while the regular accuracy, cosine similarity, rollout tests are all done with the all the transitions.

\subsection{Models}
\label{apd:models}

Our VSA-based models use embedding dimensions of $D = 512$. HRR initializes its weights as real-valued numbers sampled from a normal distribution with mean = 0 and standard deviation = 1. We utilize circular convolution for the binding and circular correlation for unbinding. For FHRR, we sample the elements from a uniform distribution from -pi to pi and utilize element-wise complex multiplication for binding. For MLP-based models, we use $D=64$ and $D =16$ for the state and action respectively. These state and actions are concatenated and fed into a MLP for next state prediction. MLP-S is constructed with 2 hidden layer $(D = 128)$, MLP-M with 4 hidden layers $(D = 256)$, and MLP-L with 6 hidden layers $(D = 512)$. Every hidden layer's output passes through a ReLU activation as well.

\subsection{Training Objectives and Hyperparameters}\label{apd:objective}
For both VSA and MLP-based models. we utilize MSE for the binding loss in~\ref{binding_loss}. For all experiments, the binding
$\lambda_{\text{bind}} = 2$, $\lambda_{\text{inv}} = 0.5$, and $\lambda_{\text{ortho}} = 0.05$. The learning rate for VSA models is set as $0.007$ where as the learning rate for MLP-based models is set as $0.0005$. We apply a gradient clipping of 1 to help with learning as well.

\subsection{Experiments}\label{apd:experiment}
All experiments were run conducted over 500 epochs. For the rollout tests, we sample 500 trials of random trajectories based on the horizon length (i.e. Rollout Length = 20 implies a random trajectory with $t = 20$). We apply the cleanup operation to both VSA and MLP-based models every 2 time steps when specified as rollout + cleanup. 

\begin{figure}
    \centering
    \includegraphics[width=\linewidth]{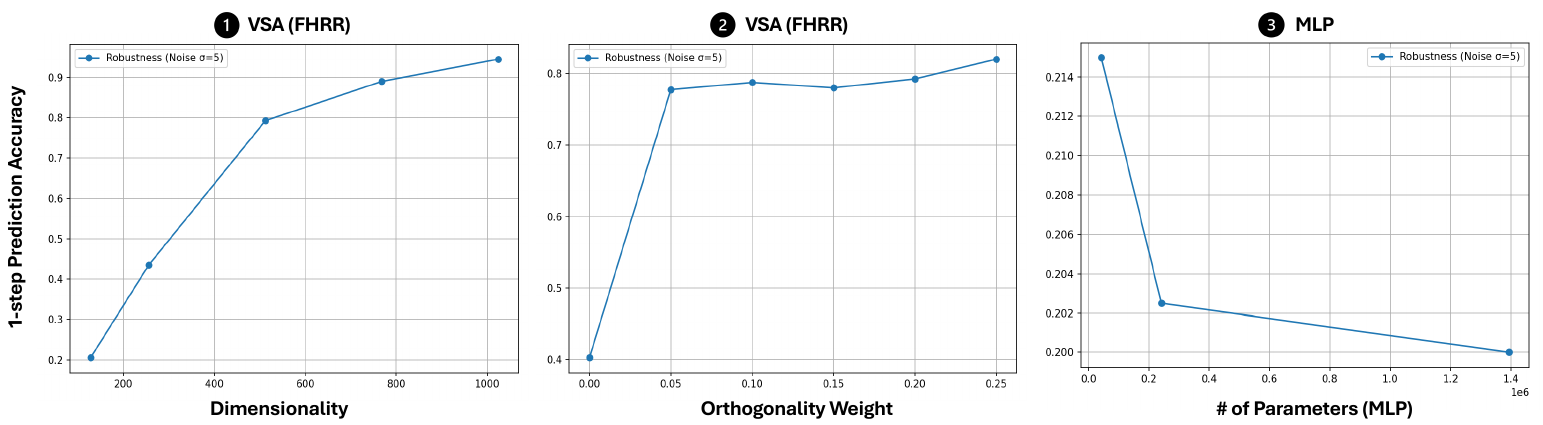}
    \caption{Robustness Ablation Study}
    \label{fig:robustness_ablation}
\end{figure}

In~\figureref{fig:robustness_ablation} we run an ablation study and see that increasing dimensionality helps with robustness as expected. Having an orthgonality weight is also necessary but shows little benefit of increasing the weight itself. On the other hand, increasing the number of parameters hurts the robustness of an MLP-based model.

\subsection{Inference Details}\label{apd:inference}

In~\tableref{tbl:HDC_INFER_Results}, we display the number of parameters of each model as well as the inference time. Experiments were conducted using an NVIDIA GPU 3060 Ti and inference times were reported in milliseconds.

\begin{table}[hbtp]
\label{tbl:HDC_INFER_Results}
\setlength{\tabcolsep}{3pt}
\floatconts
  {tab:example-booktabs}
  {\caption{VSA vs MLP Parameter and Inference Speed}}
  {\begin{tabular}{lccccc}
    \toprule
     & VSA (HRR) & VSA (FHRR) & MLP-S &  MLP-M & MLP-L \\
    \midrule
    Parameter Count & 53,248 & 53,248 & 41,600 & 241,024 & 1,394,048 \\
    Parameter Ratio & 1x & 1x & 0.8x & 4.5x & 26.2x \\
    Inference time (ms) & 0.2063 & 0.1528 & 0.1174 & 0.1715 & 0.3135 \\
    Inference + Clean time (ms) & 0.2632 & 0.2421 & 0.1743 & 0.2317 & 0.3761 \\
    \bottomrule
    \end{tabular}}
\end{table}

\end{document}